%% file: main.tex
\documentclass{llncs}

\usepackage{hyperref}
\usepackage[framemethod=TikZ]{mdframed}
\usepackage{multicol}
\usepackage{xcolor}
\usepackage{comment}
\usepackage[final]{changes}
\usepackage{caption}
\usepackage{graphicx}
\usepackage{url}
\usepackage{todonotes}
\usepackage{amsmath}
\usepackage{amssymb}
\usepackage{bm}
\usepackage{multirow}
\usepackage{amsfonts}
\usepackage{enumitem}
\usepackage{tcolorbox}
\usepackage{booktabs}
\usepackage{tabularx}
\usepackage{rotating}
\usepackage{marginnote}
\usepackage{subfig}

\begin{document}

\title{Alarm-Based Prescriptive Process Monitoring\thanks{Work supported by the European Community's FP7 Framework Program under grant n. 603993 (CORE) and by the Estonian Research Council (IUT20-55)}
}
\author{Irene Teinemaa\inst{1} \and Niek Tax\inst{2} \and Massimiliano de Leoni\inst{2} \and \\ Marlon Dumas\inst{1} \and Fabrizio Maria Maggi\inst{1}}

\institute{University of Tartu, Estonia,\\ \email{\{irene.teinemaa,marlon.dumas,f.m.maggi\}@ut.ee} \and
Eindhoven University of Technology, The Netherlands\\
\email{\{n.tax,m.d.leoni\}@tue.nl}}

\maketitle

\begin{abstract}
Predictive process monitoring is concerned with the analysis
of events produced during the execution of a process in order to predict the future state of ongoing cases thereof. Existing techniques in this field are able to predict, at each step of a case, the likelihood that the case will end up in an undesired outcome. These techniques, however, do not take into account what process workers may do with the generated predictions in order to decrease the likelihood of undesired outcomes. This paper proposes a framework for prescriptive process monitoring, which extends predictive process monitoring approaches with the concepts of alarms, interventions, compensations, and mitigation effects. The framework incorporates a parameterized cost model to assess the cost-benefit tradeoffs of applying prescriptive process monitoring in a given setting. The paper also outlines an approach to optimize the generation of alarms given a dataset and a set of cost model parameters. The proposed approach is empirically evaluated using a range of real-life event logs.
\end{abstract}

\input{introduction}
\input{background}
\input{framework}

\input{approach}

\input{evaluation}

\input{related_work}
\input{threats}
\input{conclusion}

\bibliographystyle{splncs}
\bibliography{bibliography,maxBib}

\end{document}

%% file: introduction.tex
\section{Introduction}
\label{sec:intro}

\emph{Predictive process monitoring}~\cite{maggi2014predictive,MetzgerLISFCDP15} is a family of techniques to predict the future state of ongoing cases of a business process based on event logs recording past executions thereof.
A predictive process monitoring technique may provide predictions on the remaining execution time of an ongoing case, the next activity to be executed, or the final outcome of the case wrt.\ a set of possible outcomes. This paper is concerned with the latter type of predictive process monitoring, which we call \emph{outcome-oriented}~\cite{teinemaa2017outcome}. For example, in a lead-to-order process, an outcome-oriented predictive process monitoring technique may predict whether a case will end up in a purchase order (desired outcome) or not (undesired outcome).





Existing outcome-oriented predictive process monitoring techniques are able to predict, after each event of a case, the likelihood that the case will end up in an undesired outcome. These techniques are restricted in scope to predicting. They do not suggest nor prescribe how and when process workers should intervene in order to decrease the likelihood of undesired outcomes.

This paper proposes a framework to extend outcome-oriented predictive process monitoring techniques in order to make them prescriptive.
Concretely, the proposed framework extends a given outcome-oriented predictive process monitoring model with a mechanism for generating alarms that lead to interventions, which, in turn, mitigate (or altogether prevent) undesired outcomes.
The proposed framework is armed with a parameterized cost model that captures, among others, the tradeoff between the cost of an intervention and the cost of an undesired outcome.
Based on this cost model, the paper outlines an approach for return on investment analysis of a prescriptive process monitoring system under a configuration of cost parameters and a predictive model trained on a given dataset.
Finally, the paper proposes and empirically evaluates an approach to tune the generation of alarms to minimize the expected cost for a given dataset and set of parameters.


The paper is structured as follows. Section~\ref{sec:background} introduces basic concepts and notations. Next, Section~\ref{sec:framework} presents the prescriptive process monitoring framework, Section~\ref{sec:approach} outlines the approach to optimize the alarm generation mechanism, and Section~\ref{sec:evaluation} reports on our empirical evaluation. Finally, Section~\ref{sec:related} discusses related work, Section~\ref{sec:threats} delineates the limitations of our framework and consequent future work, and Section~\ref{sec:conclusion} summarizes the contributions.

%% file: background.tex
\section{Background: Events, Traces, and Event Logs}
\label{sec:background}
For a given set $A$, $A^*$ denotes the set of all sequences over $A$ and $\sigma=\langle a_1,a_2,\dots,a_n\rangle$ a sequence of length $n$; $\langle\rangle$ is the empty sequence and $\sigma_1 \cdot \sigma_2$ is the concatenation of sequences $\sigma_1$ and $\sigma_2$. $\mathit{hd}^k(\sigma)=\langle a_1, a_2, \dots, a_k\rangle$ is the prefix of length $k$ ($0 < k < n$) of sequence $\sigma$. For example, $\mathit{hd}^2(\langle a,b,c,d,e\rangle)=\langle a,b\rangle$.


Let $\mathcal{E}$ be the event universe, i.e., the set of all possible event identifiers, and $\mathcal{T}$ the time domain. We assume that events are characterized by various properties, e.g., an event has a timestamp, corresponds to an activity, is performed by a particular resource, etc. We do not impose a specific set of properties, however, we assume that two of these properties are the timestamp and the activity of an event, i.e., there is a function $\pi_\mathcal{T}\in \mathcal{E}\rightarrow\mathcal{T}$ that assigns timestamps to events, and a function $\pi_\mathcal{A}\in\mathcal{E}\rightarrow\mathcal{A}$ that assigns to each event an activity from a finite set of process activities $\mathcal{A}$. An \emph{event log} is a set of events, each linked to one trace and globally unique, i.e., the same event cannot occur twice in a log. A trace in a log represents the execution of one case.

\begin{definition}[Trace, Event Log]
	A \emph{trace} is a finite non-empty sequence of events $\sigma\in\mathcal{E}^*$ such that each event appears only once and time is non-decreasing, i.e., for $1\le i < j \le |\sigma|:\sigma(i)\neq\sigma(j)$ and $\pi_\mathcal{T}(\sigma(i))\le\pi_\mathcal{T}(\sigma(j))$. An \emph{event log} is a set of traces $L\subset\mathcal{E}^*$ such that each event appears at most once in the entire log.\looseness=-1
\end{definition}



	


%% file: framework.tex
\newcommand{\true}{\textsf{true}}
\newcommand{\false}{\textsf{false}}
\newcommand{\avg}{\textsf{avg}}

\newcounter{framecnt}

\newenvironment{scenario}[1]
    {\begin{figure}[t!]
    \refstepcounter{framecnt}
    \renewcommand{\theHfigure}{cont.\arabic{framecnt}}
    \begin{mdframed}[roundcorner=10pt,backgroundcolor=blue!10]
    \textbf{\centerline{Box \arabic{framecnt} --- Scenario ``#1''}}
    \smallbreak

    \begin{scriptsize}
    }
    {\end{scriptsize}
    \end{mdframed}
     \end{figure}
    }

%
%
%

\section{Prescriptive Process Monitoring Framework}
\label{sec:framework}

In this section, we introduce a cost model for alarm-based prescriptive process monitoring  and illustrate this model using three scenarios (Section~\ref{sec:costs_model}). We then formalize the concept of alarm system (Section~\ref{sec:costInstances})  and discuss conditions under which an alarm system has a positive return on investment (Section~\ref{sec:roi}).

\subsection{Concepts and Cost Model}
\label{sec:costs_model}
An alarm-based prescriptive process monitoring system (\emph{alarm system} for short) is a monitoring system that raises an alarm in relation to a running case of a business process, in order to indicate that the case is likely to lead to an undesired outcome. These alarms are handled by process workers who intervene by performing an action (e.g., calling a customer or blocking a credit card) in order to prevent or mitigate the undesired outcome. These actions may have a cost, which we call \emph{cost of intervention}. Instead, if the case ends in a negative outcome, this leads to a cost called \emph{cost of undesired outcome}.




As an example, consider a municipality that needs to collect city taxes. If the inhabitants do not pay their taxes on time, the municipality may run into cash flow issues. Accordingly, in case of an unpaid tax debt (undesired outcome), the municipality may decide to outsource the debt collection to an external collection agency, for which it has to pay a recovery fee. These fees constitute the cost of the undesired outcome.
In light of their characteristics and past payment history, certain inhabitants may have a higher risk of missing the payment deadline. Therefore, sending a reminder letter to these high-risk inhabitants may increase the likelihood of receiving the payment on time. However, such an intervention comes with costs related to preparing the letter by an employee (proportional to the employee's hourly salary rate) and the postal costs for sending the letter.

In certain scenarios, the cost of an intervention may increase over time, acknowledging the importance of alarming as early as possible. For instance, in a railway maintenance process, if an alarm about a possible railway disruption is raised early, the problem could be solved with regular maintenance procedures. Conversely, if the alarm is raised when the need for maintenance has become urgent, the maintenance provider could be required to allocate more resources in order to solve the problem on time.

When an alarm is raised, there is a certain probability, but no certainty, that the case will reach an undesired outcome if no intervention is made. If the case does not conclude with an undesired outcome even without interventions, doing the intervention causes unnecessary costs (e.g., a company could lose customers and/or opportunities). The cost related to such unnecessary interventions is referred to as \emph{cost of compensation}.
For instance, financial institutions may block credit card payments when they suspect that a card was cloned. However, in some cases, it may happen that the suspicion was unfounded and that the payment was legitimate. If these cases become too frequent, the reputation of the financial institution could be hampered.\looseness=-1

The purpose of alarming is to avoid an undesired outcome. However, in several scenarios, it is not possible to
fully prevent the cost of the undesired outcome, while the intervention could still help to mitigate it. Based on this rationale, we introduce the concept of \emph{mitigation effectiveness} of an intervention, reflecting the proportion of the cost of an undesired outcome that can be avoided by carrying out the intervention. Oftentimes, the mitigation effectiveness decreases with time, i.e., the earlier the intervention takes place, the higher is the proportion of costs that can be avoided.
Consider, for instance, the process of paying unemployment benefits by a social security institution.
In this case, the aim of an alarm system could be to notify the institution about citizens who might be receiving unentitled benefits. Since the benefits that have already been issued are unlikely to be recollected, the cost of the undesired outcome cannot be avoided completely. Therefore, it is important to raise the alarm as early as possible, in order to effectively mitigate the cost of the undesired outcome.

An alarm system is intended as a system where cases are continuously monitored. However, since continuous monitoring is impractical, we assume that cases are monitored after each executed event and, therefore, alarms can only be raised after that an event has occurred.
In the remainder, each case is identified by a trace $\sigma$ that is (eventually) recorded in an event log.
Definition~\ref{def:costs} formalizes the costs defined above. Since costs may depend on the position in the case in which the alarm is raised and/or on other cases being executed, we define the costs as functions over the number of already executed events and over the entire set of cases under execution.
\begin{definition}[Alarm-based Cost Model]
\label{def:costs}
An \emph{alarm-based cost model} is a tuple $(c_\mathit{in},c_\mathit{out},c_\mathit{com},\mathit{eff})$ consisting of:
\begin{itemize}[noitemsep,topsep=0pt]
\item a function $c_\mathit{in}\in\mathbb{N}\times\mathcal{E}^*\times 2^{\mathcal{E}^*}\rightarrow \mathbb{R}^+_0$ modeling the \emph{cost of \underline{in}tervention}:
given a trace $\sigma$ belonging to an event log $L$, $c_\mathit{in}(k,\sigma,L)$ indicates the cost of an intervention in $\sigma$ when the intervention takes place after the $k$-th event;

\item a function $c_\mathit{out}\in\mathcal{E}^*\times 2^{\mathcal{E}^*}\rightarrow \mathbb{R}^+_0$ modeling the \emph{cost of undesired \underline{out}come};

\item a function $c_\mathit{com}\in\mathcal{E}^*\times 2^{\mathcal{E}^*}\rightarrow \mathbb{R}^+_0$ modeling the \emph{cost of \underline{com}pensation};

\item a function $\mathit{eff} \in\mathbb{N}\times\mathcal{E}^*\times 2^{\mathcal{E}^*}\rightarrow[0,1]$ modeling the \emph{mitigation \underline{eff}ectiveness} of an intervention:
given a trace $\sigma$ belonging to an event log $L$, $\mathit{eff}(k,\sigma,L)$ indicates the mitigation effectiveness of an intervention in $\sigma$ when the intervention takes place after the $k$-th event.
\end{itemize}
\end{definition}
\begin{scenario}{Unemployment Benefits}\label{ex:UWV}
In several countries, a social security institution is responsible for the execution of a number of employee-related insurances, such as unemployment benefits.
When residents (hereafter customers) become unemployed, they are usually entitled to monthly monetary benefits for a certain period of time.
These payments are stopped when the customer reports that he/she has found a new job. Unfortunately, several customers omit to inform the institution about finding a job and, thus, keep receiving benefits they are not entitled to. Those customers are expected to return the amount of benefits that they have received unlawfully. However, in practice, this rarely happens and the overpaid amount is lost to the institution.
In light of the above, the social security institution would benefit from an alarm system that would inform about customers who are likely to be receiving unentitled benefits.
Let $\mathit{unt}(\sigma)$ denote the amount of unentitled benefits received in a case corresponding to trace $\sigma$.
Based on discussions with the stakeholders of a real social security institution, we designed the following cost model instantiation for such an alarm system.
\begin{description}
  \item[Cost of intervention.] For the intervention, an employee needs to check if the customer is indeed receiving unentitled benefits and, if so, fill in the forms for stopping the payments. Let $S$ be the employee's average salary rate per time unit; let $i_s$ and $i_f$ denote the positions of the events in $\sigma$ when the employee started working on the intervention and finished it, respectively. The cost of an intervention can be modeled as: $c_\mathit{out}(\sigma,L)=(\pi_\mathcal{T}(\sigma(i_f))-\pi_\mathcal{T}(\sigma(i_s)))\cdot S$.
  \item[Cost of undesired outcome.] The total amount of unentitled benefits that the customer would obtain without stopping the payments, i.e.,
  $c_\mathit{out}(\sigma,L)=\mathit{unt}(\sigma)$.
  \item[Cost of compensation.] The social security institution works in a situation of monopoly, which means that the customer cannot be lost because of moving to a competitor, i.e., there is no cost of compensation: $c_\mathit{com}(\sigma,L)=0$.
  \item[Mitigation effectiveness.] The proportion of unentitled benefits that will not be paid thanks to the intervention: $\mathit{eff}(k,\sigma,L)=\frac{\mathit{unt}(\sigma)-\mathit{unt}(\mathit{hd}^k(\sigma))}{\mathit{unt}(\sigma)}$. Note that this cost function is not employed if there is no undesired outcome (i.e., if $\mathit{unt}(\sigma)=0$).
\end{description}
\end{scenario}
\begin{scenario}{Financial Institution}\label{ex:bank}
Suppose that the customers of a financial institution use their credit cards to make payments online. Each such transaction is associated with a risk that the transaction is made through a cloned card. In this scenario, an alarm system is intended to determine whether the credit card needs to be blocked due to a high risk of being cloned. However, in case the credit card is not malicious, blocking the card would cause discomfort to the customer who may consequently opt to switch to a different financial institution.
Let $\sigma$ be the trace of credit card transactions for a customer and $\mathit{value}(\sigma)$ the total amount of money related to malicious transactions in $\sigma$, the following is a possible cost model instantiation for this scenario.
\begin{description}
  \item[Cost of intervention.] The card is automatically blocked by the system and, therefore, the intervention costs are limited to \textsc{Post\_Cost}, i.e., to the costs for sending a new credit card to the customer by mail: $c_\mathit{in}(k,\sigma,L)=$ \textsc{Post\_Cost}.
  \item[Cost of undesired outcome.] The total amount of money related to malicious transactions that the bank would need to reimburse to the legitimate customer:
  $c_\mathit{out}(\sigma,L)=\mathit{value}(\sigma)$.
  \item[Cost of compensation.] Denoting the asset value of a customer (consisting of the amount of the investment portfolio, the account balance, etc.) with $\mathit{asset}(\sigma)$ and supposing that a fraction $p$ (i.e., $p \in [0,1]$) of the customers would switch to a different institution, the cost of compensation can be estimated as the value of the lost asset (the customer), multiplied by $p$: $c_\mathit{com}=p\cdot\mathit{asset}(\sigma)$.
  \item[Mitigation effectiveness.] The proportion of the total amount of money related to malicious transactions that does not need to be reimbursed by blocking the credit card after that $k$ events have been executed: $\mathit{eff}(k,\sigma,L)=\frac{\mathit{value}(\sigma)-\mathit{value}(\mathit{hd}^k(\sigma))}{\mathit{value}(\sigma)}$.
\end{description}
\end{scenario}
\begin{scenario}{Railway Maintenance}\label{ex:railway}
In a process for railway maintenance, an alarm should be raised when there is a risk that the railway may break down within a relatively short time range. Railway breakdowns can cause severe disruptions in the train transportation (i.e., trains could be canceled or delayed), thereby causing losses of reimbursing tickets to travelers.
\begin{description}
  \item[Cost of intervention.] The cost of an intervention increases with time because the more urgent the disruption is, the more resources need to be allocated for handling it. We assume that the cost is at its minimum $m$ at the beginning of a trace $\sigma$ and grows exponentially with time: $c_\mathit{in}(k,\sigma,L)=m \cdot \beta\exp(\pi_\mathcal{T}(\sigma(k)))$ for some $\beta > 0$.
  \item[Cost of undesired outcome.] Let $P$ be the average total price of tickets sold per time unit; let $i_d(\sigma)$ and $i_m(\sigma)$ be the positions of the events in $\sigma$ when the disruption took place and was resolved, respectively. The cost of the undesired outcome can be calculated as $P$ multiplied by the length of the timeframe when the railway service was disrupted: $c_\mathit{out}(\sigma,L)=(\pi_\mathcal{T}(\sigma(i_m))-\pi_\mathcal{T}(\sigma(i_d)))\cdot P$.
  \item[Cost of compensation.] Assuming that performing (unnecessary) maintenance actions does not cause inconveniences to the customers, no cost of compensation is present: $c_\mathit{com}(\sigma,L)=0$.
  \item[Mitigation effectiveness.] A timely intervention fully avoids the undesired outcome: $\mathit{eff}(k,\sigma,L)=1$ for any $k \in [1,|\sigma|]$.
\end{description}
\end{scenario}

To illustrate the versatility of the above cost model, we discuss three use cases for alarm systems and their corresponding cost model configurations. The first scenario, in Box 1, refers to the provision of unemployment benefits. The cost model for this scenario is based on several discussions with the stakeholders of a real social security institution~\cite{D_dL_M@COOP17}.
The second scenario, in Box 2, refers to the detection of malicious credit card payments in a financial institution. Differently from the previous scenario, in this case, there is a risk of cost of compensation: due to the inconvenience caused by blocking their credit card, customers can switch to competitors.
Box 3 refers to the process of predictive maintenance in railway services.
This scenario is different from the previous ones because, in this case, the cost of an intervention increases over time.

%

\subsection{Alarm-Based Prescriptive Process Monitoring System}
\label{sec:costInstances}
An alarm-based prescriptive process monitoring system is driven by the outcome of the cases. Hereon, the outcome of the cases is represented by a function $\mathit{out} \in \mathcal{E}^*\rightarrow \{\true,\false\}$: given a case identified by a trace $\sigma$, if the case has an undesired outcome, $\mathit{out}(\sigma)=\true$; otherwise, $\mathit{out}(\sigma)=\false$. In reality, during the execution of a case, its outcome is not yet known and needs to be estimated based on past executions that are recorded in an event log $L \subset \mathcal{E}^*$. The outcome estimator is a function $\widehat{out}_L\in\mathcal{E}^*\rightarrow[0,1]$ predicting the likelihood $\widehat{out}_L(\sigma')$ that the outcome of a case that starts with prefix $\sigma'$ is undesired. We can define an alarm system as a function that returns true or false depending on whether an alarm is raised based on the predicted outcome or not.
\begin{definition}[Alarm-Based Prescriptive Process Monitoring System]
\label{def:alarm}
Given an event log $L \subset \mathcal{E}^*$, let $\widehat{out}_L$ be an outcome estimator built from $L$.
An \emph{alarm-based prescriptive process monitoring system} is a function $\mathit{alarm}_{\widehat{out}_L} \in \mathcal{E}^*\rightarrow \{\true,\false\}$.
Given a running case identified by a trace $\sigma$ and with current prefix $\sigma'$, $\mathit{alarm}_{\widehat{out}_L}(\sigma)$ returns $\true$, if an alarm is raised based on the predicted outcome $\widehat{out}_L(\sigma')$, or $\false$, otherwise.
\end{definition}
For simplicity, we omit the subscript $L$ from $\widehat{out}_L$ and omit $\widehat{out}_L$ from $\mathit{alarm}_{\widehat{out}_L}$ when it is clear from the context. An alarm system can raise an alarm at most once per case, since we assume that already the first alarm triggers an intervention by the stakeholders.
\looseness=-1

\begin{table}[tb]
	\centering
	\vspace{-0.2cm}
\caption{Cost of a case $\sigma$ based on its outcome and whether an alarm was raised}		
\label{table:cost_matrix_business_costs}
	\begin{tabular}{c|c|c}
		\toprule
		& undesired outcome & desired outcome \\
		\midrule
		alarm raised & $c_\mathit{in}(k,\sigma,L) + (1-\mathit{eff}(k,\sigma,L)) c_\mathit{out}(\sigma,L)$ & $c_\mathit{in}(k,\sigma,L) + c_\mathit{com}(\sigma,L)$ \\
		alarm not raised & $c_\mathit{out}(\sigma,L)$ & $0$ \\
		\bottomrule
	\end{tabular}
\vspace{-0.3cm}
\end{table}

The purpose of an alarm system is to minimize the cost of executing a case. \tablename~\ref{table:cost_matrix_business_costs} summarizes how the cost of a case is determined based on a cost model (cf.\ Def.~\ref{def:costs}), on the case outcome, and on whether an alarm was raised or not.
\begin{definition}[Cost of Case Execution]
\label{sec:instanceExecution}
Let $cm=(c_\mathit{in},c_\mathit{out},c_\mathit{com},\mathit{eff})$ be an alarm-based cost model.
Let $out\in \mathcal{E}^*\rightarrow \{\true,\false\}$ be an outcome function.
Let $\mathit{alarm} \in \mathcal{E}^*\rightarrow \{\true,\false\}$ be an alarm-based prescriptive process monitoring system.
Let $L \subset E^*$ be the entire set of \emph{complete} (i.e., no more running) cases.
Let $\sigma \in L$ be a case.
Let $\mathcal{I}(\sigma,\mathit{alarm})$ be the index of the event in $\sigma$ when the alarm was raised or zero if no alarm was raised:

\noindent\resizebox{\linewidth}{!}{
$\mathcal{I}(\sigma,\mathit{alarm})=\begin{cases}
0& \text{if }\forall{k\in[1,|\sigma|{-}1]}. \neg \mathit{alarm}(\mathit{hd}^k(\sigma)),\\
1 & \text{if } \mathit{alarm}(\mathit{hd}^1(\sigma)),\\
i \in [2,|\sigma|] \text{ s.t. } \mathit{alarm}(\mathit{hd}^i(\sigma)) \land & \text{otherwise.}\\
\quad \forall k\in[1,i-1].\; \neg \mathit{alarm}(\mathit{hd}^k(\sigma)) &
\end{cases}$}

\noindent The \emph{cost of execution of case $\sigma$} supported by the alarm system is:

\noindent\resizebox{\linewidth}{!}{
$    \mathit{cost}(\sigma,L,cm,\mathit{alarm})=
\begin{cases}
c_\mathit{in}(\mathcal{I}(\sigma,\mathit{alarm}),\sigma,L) + (1-\mathit{eff}(\mathcal{I}(\sigma,\mathit{alarm}),\sigma,L))\cdot c_\mathit{out}(\sigma,L)& \mathit{out}(\sigma)\land \mathcal{I}(\sigma,\mathit{alarm})>0, \\
c_\mathit{in}(\mathcal{I}(\sigma,\mathit{alarm}),\sigma,L)+c_\mathit{com}(\sigma,L)              & \neg\mathit{out}(\sigma)\land \mathcal{I}(\sigma,\mathit{alarm})>0,\\
c_\mathit{out}(\sigma,L) & \mathit{out}(\sigma)\land \mathcal{I}(\sigma,\mathit{alarm})=0,\\
0              & \text{otherwise.}
\end{cases}$
}
\end{definition}
Section~\ref{sec:approach} illustrates how an alarm-based prescriptive process monitoring system can be designed aiming at the minimization of the case execution costs (according to Def.~\ref{sec:instanceExecution}).\looseness=-1

\input{framework_roi_analysis}

%% file: framework_roi_analysis.tex
\subsection{Return on Investment Analysis}
\label{sec:roi}
In this section, we provide an analysis and guidelines that suggest when it is valuable to invest in developing an alarm system, namely, when the return on investment (ROI) is positive. To this aim, we need to compare the case of a business process execution supported by an alarm system with the \emph{as-is} situation where the business process is executed without this support. For this analysis, we consider a set of cases recorded in an event log $L$, where no interventions were done, and a cost model $cm=(c_\mathit{in},c_\mathit{out},c_\mathit{com},\mathit{eff})$. 

The \emph{as-is} situation implies that no interventions are done in any of the cases $\sigma \in L$ that lead to an undesired outcome, yielding a cost $c_\mathit{out}(\sigma)$. When applied to the entire log $L$, the cost is $\mathit{cost}_{as\textit{-}is}(L)=\sum_{\sigma \in L \text{ s.t. } \mathit{out}(\sigma)} c_\mathit{out}(\sigma)$. Instead, when a certain system $\mathit{alarm}$ is in effect, the costs are $\mathit{cost}_{\mathit{alarm}}(L)=\sum_{\sigma \in L} \mathit{cost}(\sigma,L,\mathit{cm},\mathit{alarm})$ (cf.\
 Defs.~\ref{def:costs}, \ref{def:alarm}).
With this setting, the return on investment of the system $\mathit{alarm}$ is $\mathit{ROI}(L,\mathit{cm},\mathit{alarm})=\mathit{cost}_{as\textit{-}is}(L)-\mathit{cost}_{\mathit{alarm}}(L)$, which must be positive to make deploying the system worthwhile.

The question that remains is: \emph{how does the ROI depend on the cost model and the alarm system?}
For the sake of simplicity, we assume, in this analysis, that every component of the cost model is constant. Furthermore, the initial investment costs are not considered because we assume the system to be fully operational already for a sufficiently long time, so that the the initial costs have been amortized. The above assumptions yield the following case cost:

	$    \mathit{cost}(\sigma,L,cm,\mathit{alarm})=
	\begin{cases}
	c_\mathit{in} + (1-\mathit{eff})c_\mathit{out}& \mathit{out}(\sigma)\land\mathcal{I}(\sigma,\mathit{alarm})>0,\\
	c_\mathit{in} + c_\mathit{com}             & \neg\mathit{out}(\sigma)\land\mathcal{I}(\sigma,\mathit{alarm})>0,\\
	c_\mathit{out} & \mathit{out}(\sigma)\land\mathcal{I}(\sigma,\mathit{alarm})=0,\\
	0              & \text{otherwise}
	\end{cases}$
where $c_\mathit{in}$, $c_\mathit{out}$, $c_\mathit{com}$, and $\mathit{eff}$ are constants. In order for the ROI to be positive, it should be
$\mathit{cost}_{as\textit{-}is}(L) > \mathit{cost}(\sigma,L,cm,\mathit{alarm})$, that is:
\[
\begin{footnotesize}
\begin{array}{l}
|L_{und}|\cdot c_\mathit{out} > |L_{und\&al}|(c_\mathit{in} + (1-\mathit{eff})c_\mathit{out})+|L_{des\&al}|(c_\mathit{in} + c_\mathit{com})+|L_{und\&nal}| \cdot c_\mathit{out}
\end{array}
\end{footnotesize}
\]
where $L_{und\&al}$, $L_{des\&al}$, $L_{und\&nal}$ respectively consist of the traces in $L$ related to the cases with an \underline{und}esired outcome that would be \underline{al}armed, with a \underline{des}ired outcome that would still be \underline{al}armed, with an \underline{und}esired outcome that would \underline{n}ot be \underline{al}armed; also,
$L_{und}= L_{und\&al} \cup L_{und\&nal}$.
After simplification:
\begin{equation}\label{equ:ROIsimpl}
|L_{und\&al}|(\mathit{eff}c_\mathit{out}-c_\mathit{in} ) >|L_{des\&al}|(c_\mathit{in} + c_\mathit{com}).
\end{equation}
Because the right-hand side of Eq.~\ref{equ:ROIsimpl} is non-negative, it follows as a corollary that $\mathit{eff}c_\mathit{out} > c_\mathit{in}$ is a necessary condition for return on investment. In other words, it must be possible to avoid a cost that is higher than the cost of doing the intervention. This provides a validation of our framework: it complies with the \emph{reasonableness condition} in the cost-sensitive learning literature~\cite{elkan2001foundations}, which states that the cost of labeling an example incorrectly should always be greater than the cost of labeling it correctly.

Eq.~\ref{equ:ROIsimpl} also illustrates that the policy of always alarming does not yield a positive ROI, unless the number of cases with undesired outcome and the cost of the undesired outcome are sufficiently high. When the number of cases with an undesired outcome is small (e.g., the unemployment benefits and the financial institution scenarios described in Boxes~1 and 2) and at the same time the cost of this undesired outcome is small, then the left-hand side of Eq.~\ref{equ:ROIsimpl} is negligible, thus leading to condition $c_\mathit{in} + c_\mathit{com}<0$, which can never hold.

So far we have assumed, for the sake of simplicity, that costs and mitigation effectiveness are constant, similarly to traditional cost-sensitive learning.
However, the novelty of our formulation lays in the fact that costs are functions that depend on the time when an intervention is made. As a result, the reasonableness of the cost matrix would not be fixed, but potentially changes over time.
Still, variable costs do not invalidate the ROI analysis.
In fact, in order for the ROI to be positive, it is sufficient that the cost model is reasonable for a certain time period; otherwise, the alarm system would never raise alarms because of the cost model. Clearly, the longer the reasonable-cost period is, the higher the ROI.

%% file: approach.tex
\section{Alarming Mechanisms and Empirical Thresholding}
\label{sec:approach}
An alarm system needs two components to minimize the costs of future cases: (1) a probabilistic classifier $\widehat{\mathit{out}}_L\in\mathcal{E}^*\rightarrow[0,1]$ that estimates the likelihood of an undesired outcome for a partial trace based on some historical observations $L$, and (2) an alarming mechanism that, for a given incomplete case, decides whether or not to raise an alarm based on the prediction made by $\widehat{\mathit{out}}_L$. We propose to implement the second component using a function $\mathit{agent}\in[0,1]\rightarrow\{\true,\false\}$ that operates on the estimated likelihood of an undesired outcome, where value $\true$ represents the decision to raise an alarm. Together, the two components form an \emph{alarm system}, $\mathit{alarm}(\mathit{hd}^k(\sigma))=\mathit{agent}(\widehat{\mathit{out}}_L(\mathit{hd}^k(\sigma)))$, which makes the decision on whether or not to raise an alarm based on the observed $k$ events of trace $\sigma$.

The first component, function $\widehat{\mathit{out}}_L$, can be implemented using any classification algorithm that is naturally probabilistic, i.e., that outputs likelihood scores on a $[0,1]$-interval instead of a binary outcome. Examples of probabilistic classification algorithms include naive Bayes, logistic regression, and random forest.
The classifier is trained on historical cases recorded in a log $L_\mathit{train}$.

It is easy to see that the decision on whether or not to raise an alarm should be dependent not only on $\widehat{\mathit{out}}_{L_\mathit{train}}(\mathit{hd}^k(\sigma))$, but also on the configuration of $c_\mathit{in}$, $c_\mathit{out}$, $c_\mathit{com}$, and $\mathit{eff}$. When $c_\mathit{in}$ and $c_\mathit{com}$ are very low compared to $c_\mathit{out}$, it might be beneficial to use a lower threshold for the estimated likelihood $\widehat{\mathit{out}}_{L_\mathit{train}}(\mathit{hd}^k(\sigma))$, while one would want to be more certain that the undesired outcome will happen when $c_\mathit{in}$ or $c_\mathit{com}$ is high.

We propose to implement the second component, $\mathit{agent}$, as an \emph{alarming threshold}, i.e., a mechanism that alarms when the estimated likelihood of an undesired outcome is at least $\tau$. We define function $\mathit{alarm}_\tau(\mathit{hd}^k(\sigma))$ to be the alarming function that uses the alarming mechanism $\mathit{agent}_\tau(\widehat{\mathit{out}}_{L_\mathit{train}}(\mathit{hd}^k(\sigma))) = \widehat{\mathit{out}}_{L_\mathit{train}}(\mathit{hd}^k(\sigma)) \ge \tau$.
We aim at finding the optimal value $\overline{\tau}$ of the alarming threshold that minimizes the cost on a log $L_{\mathit{thres}}$ consisting of historical observations such that $L_{\mathit{thres}}\cap L_{\mathit{train}}=\emptyset$ with respect to a given likelihood estimator $\widehat{\mathit{out}}_{L_\mathit{train}}$ and cost model $\mathit{cm}$. The total cost of an alarming mechanism $\mathit{alarm}$ on a log $L$ is defined as $\mathit{cost}(L,\mathit{cm},\mathit{alarm})=\Sigma_{\sigma\in L}\mathit{cost}(\sigma,L,\mathit{cm},\mathit{alarm})$. Using this definition, we define $\overline{\tau} = \arg\min_{\tau\in[0,1]} \mathit{cost}(L_\mathit{thres},\mathit{cm}, \mathit{alarm}_\tau)$. Optimizing a threshold $\tau$ on a separate thresholding set is called \emph{empirical thresholding}~\cite{sheng2006thresholding} and the search for the optimal threshold $\overline{\tau}$ wrt.\ a specified cost model and log $L_{\mathit{thres}}$ can be performed using any hyperparameter optimization technique, such as Tree-structured Parzen Estimator (TPE) optimization~\cite{bergstra2011algorithms}. The resulting approach can be considered to be a form of cost-sensitive learning, since the value $\overline{\tau}$ depends on how the cost model $\mathit{cm}$ is specified.

Note that as an alternative to a single global alarming threshold $\overline{\tau}$ it is possible to optimize a separate threshold $\overline{\tau_k}$ for each prefix length $k$. We experimentally found a single global threshold $\overline{\tau}$ optimized on $L_\mathit{thres}$ to outperform separate prefix-length-dependent thresholds $\overline{\tau_k}$ optimized on $L_\mathit{thres}$, therefore we propose to use a single optimized threshold.

After creating the fully functional alarm system by training a classifier on $L_\mathit{train}$ and optimizing the alarming threshold on $L_\mathit{thres}$ for the given cost model $\mathit{cm}$, the obtained alarming function $\mathit{alarm}$ can be applied to the continuous stream of events coming from the executions of a business process, thereby reducing the processing costs of the running cases.

%% file: evaluation.tex
\section{Evaluation}
\label{sec:evaluation}

In this section, we describe the experimental setup for evaluating the proposed framework and the results of the evaluation. We address the following research questions:

\begin{enumerate}[label=RQ\arabic*,leftmargin=*]
\item Can empirical thresholding find thresholds that consistently lead to a reduction in the average processing cost for different cost model configurations?
\item Does the alarm system consistently yield a benefit over different values of the mitigation effectiveness?
\item Does the alarm system consistently yield a benefit over different values of the cost of compensation?
\end{enumerate}


\subsection{Approaches and Baselines}

We experiment with two different implementations of $\widehat{\mathit{out}}_{L_\mathit{train}}$ by using different well-known classification algorithms, namely, random forest (RF) and gradient boosted trees (GBT). Both classification algorithms have shown to be amongst the top performing classification algorithms on a variety of classification tasks~\cite{fernandez2014we,olson2017data}. We employ a single classifier approach where the features for a given prefix are obtained using the aggregation encoding~\cite{de2016general}, which has been shown to perform better than alternative encodings for event logs~\cite{teinemaa2017outcome}.

We apply the TPE optimization procedure for the alarming mechanism to find the optimal threshold $\overline{\tau}$.
We use several fixed thresholds as baselines. First, we compare with the \emph{as-is} situation in which alarms are never raised. Secondly, we compare with the baseline $\tau = 0$, allowing us to compare with the situation where alarms are always raised directly at the start of a case. Finally, we compare with $\tau = 0.5$ enabling the comparison with the cost-insensitive scenario that simply alarms when an undesired outcome is expected. The implementation of the approach and the experimental setup are openly available online.\footnote{\url{https://taxxer.github.io/AlarmBasedProcessPrediction/}}

\subsection{Datasets}

For each event log, we use all available data attributes as input to the classifier. Additionally, we extract the \emph{event number}, i.e., the index of the event in the given case, the \emph{hour, weekday, month, time since case start}, and \emph{time since last event}.
Infrequent values of categorical attributes (occurring less than 10 times in the log) are replaced with value ``other'', to avoid exploding the dimensionality. Missing attributes are imputed with the respective most recent (preceding) value of that attribute in the same trace when available, otherwise with zero.
Traces are cut before the labeling of the case becomes trivially known and are truncated at the 90th percentile of all case lengths to avoid bias from very long traces.
We use the following datasets to evaluate the alarm system:
\begin{description}
  \item[BPIC2017.]  This log records execution traces from a loan application process in a Dutch financial institution.\footnote{\url{https://doi.org/10.4121/uuid:5f3067df-f10b-45da-b98b-86ae4c7a310b}}
The event log was split into two sub-logs, denoted with \emph{bpic2017\_refused} and \emph{bpic2017\_cancelled}. In the first one, the undesired cases refer to the process executions in which the applicant has refused the final offer(s) by the financial institution and, in the second one, the undesired cases consist of those cases where the financial institution has cancelled the offer(s).
  \item[Road traffic fines.] This event log originates from the Italian local police.\footnote{\url{https://doi.org/10.4121/uuid:270fd440-1057-4fb9-89a9-b699b47990f5}}
The desired outcome is that a fine is fully paid, while in the undesired cases the fine needs to be sent for credit collection.
  \item[Unemployment.] This event log corresponds to the \emph{Unemployment Benefits} scenario (Box~\ref{ex:UWV} in Section~\ref{sec:costs_model}).
The undesired outcome is that a resident will receive more benefits than entitled, causing the need for a reclamation. Privacy constraints prevent us from making this event log publicly available.
\end{description}
Table~\ref{table:dataset_stats} describes the characteristics of the event logs used. The classes are well balanced in \emph{bpic2017\_cancelled} and \emph{traffic\_fines}, while the undesired outcome is more rare in case of \emph{unemployment} and \emph{bpic2017\_refused}. In \emph{traffic\_fines}, the traces are very short, while in the other datasets the traces are longer.\looseness=-1

\begin{table}[t]
\vspace{-0.2cm}
\caption{Dataset statistics}
\label{table:dataset_stats}
\begin{center}
\resizebox{0.8\textwidth}{!}{
	\begin{tabular}{@{}lcccccc@{}}
	\toprule
  & \# & class & min & med & (trunc.) max & \# \\
dataset name &  traces & ratio & length & length & length &  events \\ \midrule
bpic2017\_refused & 31\,413 & 0.12 & 10 & 35 & 60 & 1\,153\,398 \\
bpic2017\_cancelled & 31\,413 & 0.47 & 10 & 35 & 60 & 1\,153\,398 \\
traffic\_fines & 129\,615 & 0.46 & 2 & 4 & 5 & 445\,959 \\
unemployment & 34\,627 & 0.2 & 1 & 21 & 79 & 1\,010\,450 \\
	\bottomrule
	\end{tabular}
	}
\end{center}
\vspace{-0.2cm}
\end{table}

\subsection{Experimental Setup}
We apply a temporal split, i.e., we order the cases by their start time and from the first 80\% of the cases randomly select 80\% (i.e., 64\% of the total) for $L_\mathit{train}$ and 20\% (i.e., 16\% of the total) for $L_\mathit{thres}$, and use the remaining 20\% as the test set $L_\mathit{test}$. The events in cases in $L_\mathit{train}$ and $L_\mathit{thres}$ that overlap in time with $L_\mathit{test}$ are discarded in order to not use any information that would not be available yet in a real setting. We use TPE with 3-fold cross validation on $L_\mathit{train}$ to optimize the hyperparameters for RF and GBT.
We optimize the alarming threshold $\overline{\tau}$ by building the final classifiers using all the traces in $L_\mathit{train}$ and search for $\overline{\tau}$ using $L_\mathit{thres}$.


It is common in cost-sensitive learning to apply calibration techniques to the resulting classifier in order to obtain accurate probability estimates and, therefore, more accurate estimates of the expected cost~\cite{zadrozny2001learning}. However, we found that calibrating the classifier using Platt scaling~\cite{platt1999probabilistic} does not consistently improve the estimated likelihood of undesired outcome on the four event logs, and frequently even leads to less accurate likelihood estimates. Therefore, we decided to skip the calibration step. Moreover, since we use empirical thresholding, it is not necessary that the probabilities are well calibrated and it is sufficient that the likelihoods are reasonably ordered.

Table~\ref{table:cost_models_eval} shows the configurations of the cost model that we explore in the evaluation. To answer RQ1, we vary the ratio between $c_\mathit{out}(\sigma,L)$ and $c_\mathit{in}(k,\sigma,L)$ (keeping $c_\mathit{com}(\sigma,L)$ and $\mathit{eff}(k,\sigma,L)$ unchanged). To answer RQ2, we vary both $\mathit{eff}(k,\sigma,L)$ and the ratio between $c_\mathit{out}(\sigma,L)$ and $c_\mathit{in}(k,\sigma,L)$. To answer RQ3, we vary two ratios: 1) between $c_\mathit{out}(\sigma,L)$ and $c_\mathit{in}(k,\sigma,L)$ and 2) between $c_\mathit{in}(k,\sigma,L)$ and $c_\mathit{com}(\sigma,L)$.

\begin{table}[t]
\vspace{-0.2cm}
\caption{Cost model configurations}
\label{table:cost_models_eval}
\begin{center}
\resizebox{1\textwidth}{!}{
	\begin{tabular}{@{}lcccc@{}}
	\toprule
 & $c_\mathit{out}(\sigma,L)$ & $c_\mathit{in}(k,\sigma,L)$ & $c_\mathit{com}(\sigma,L)$ & $\mathit{eff}(k,\sigma,L)$ \\ \midrule
RQ1 & $\{1, 2, 3, 5, 10, 20\}$ & $1$ & $0$ & $1 - k / |\sigma|$ \\
RQ2 & $\{1, 2, 3, 5, 10, 20\}$ & $1$ & $0$ & $\{0, 0.1, 0.2, ..., 1\}$ \\
RQ3 & $\{1, 2, 3, 5, 10, 20\}$ & $1$ & $\{0, 1/20, 1/10, 1/5, 1/2, 1, 2, 5, 10, 20\}$ & $1 - k / |\sigma|$ \\
	\bottomrule
	\end{tabular}
	}
\end{center}
\vspace{-0.2cm}
\end{table}

We measure the average processing cost per case in $L_\mathit{test}$, and aim at minimizing this cost.
Additionally, we measure the \emph{benefit} of the alarm system, i.e., the reduction in the average processing cost of a case when using the alarm system compared to the average processing cost when not using it.



\subsection{Results}

\begin{figure}[t]
\vspace{-0.2cm}
\centering
\includegraphics[width=1\textwidth]{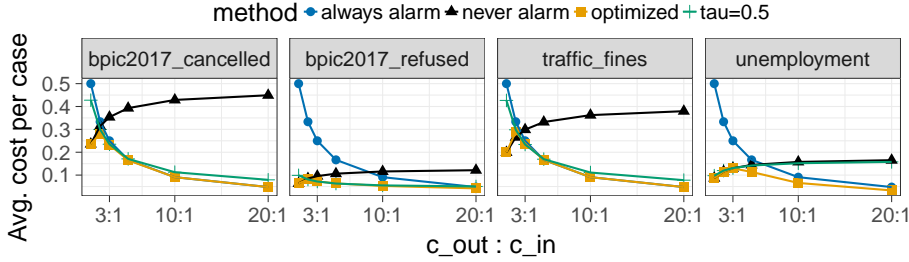}
\caption{Cost over different ratios of $c_\mathit{out}(\sigma,L)$ and $c_\mathit{in}(k,\sigma,L)$ (GBT)}
\label{fig:results_ratios}
\vspace{-0.2cm}
\end{figure}

\figurename~\ref{fig:results_ratios} shows the average cost per case when increasing the ratio of $c_\mathit{out}(\sigma,L)$ and $c_\mathit{in}(k,\sigma,L)$ from left to right. We only present the results obtained with GBT as we found it to slightly outperform RF. When the ratio between these two costs is balanced (i.e., 1:1), the minimal cost is obtained by never alarming. This is in agreement with the ROI analysis, where we found $\mathit{eff}c_\mathit{out}>c_\mathit{in}$ to be a necessary condition for having an advantage from an alarm system.
When $c_\mathit{out} \gg c_\mathit{in}$ the best strategy is to always alarm. When $c_\mathit{out}$ is slightly higher than $c_\mathit{in}$ the best strategy is to sometimes alarm based on $\widehat{out}$.
We found that the optimized $\overline{\tau}$ always outperforms the baselines. An exception is ratio 2:1 for \emph{traffic\_fines}, where never alarming is slightly better. 

\begin{figure}[t]
\vspace{-0.2cm}
\centering
\includegraphics[width=1\textwidth]{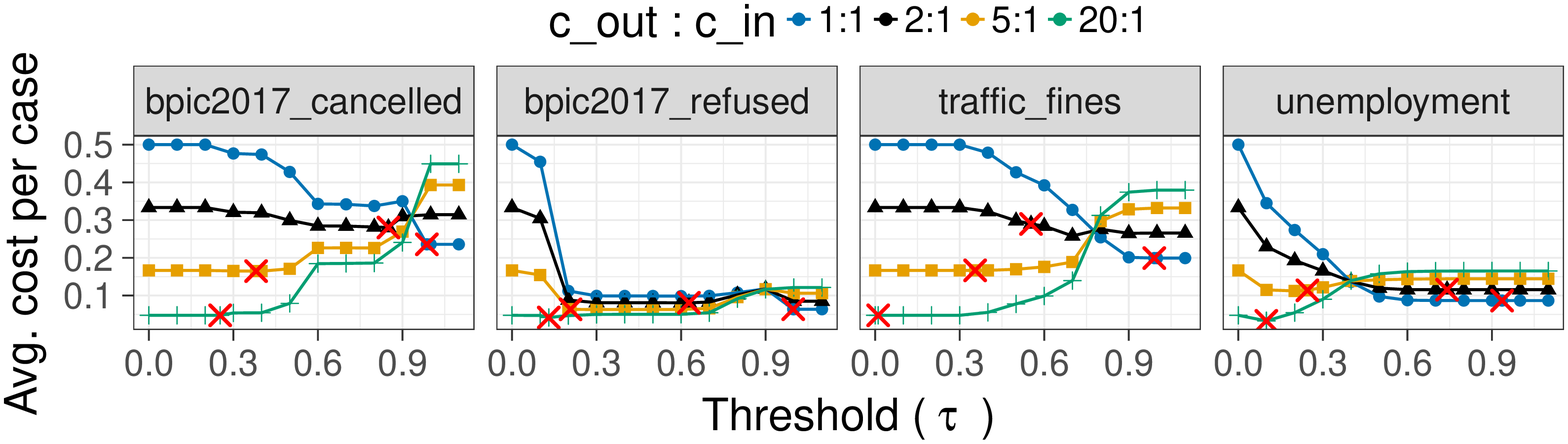}
\caption{Cost over different thresholds ($\overline{\tau}$ is marked with a red cross)}
\label{fig:results_thresholds}
\vspace{-0.2cm}
\end{figure}

In \figurename~\ref{fig:results_thresholds}, the average cost per case is plotted against different (fixed) thresholds. The optimized threshold is marked with a red cross and each line represents one particular cost ratio.
We observe that, while the optimized threshold generally obtains minimal costs, there sometimes exist multiple optimal thresholds for a given cost model configuration. For instance, in the case of the 5:1 ratio in \emph{bpic2017\_cancelled}, all thresholds between 0 and 0.4 are cost-wise equivalent. We conclude that the empirical thresholding approach consistently finds a threshold that yields the lowest cost in a given event log and cost model configuration (cf.~RQ1).

\begin{figure}[t]
\vspace{-0.2cm}
 \label{fig:results_heatmaps}
		\hspace{-0.25cm}
        \subfloat[Varying $\mathit{eff}(k,\sigma,L)$\label{fig:results_effectiveness_const_lgbm_selected}]{\includegraphics[width=0.515\linewidth]{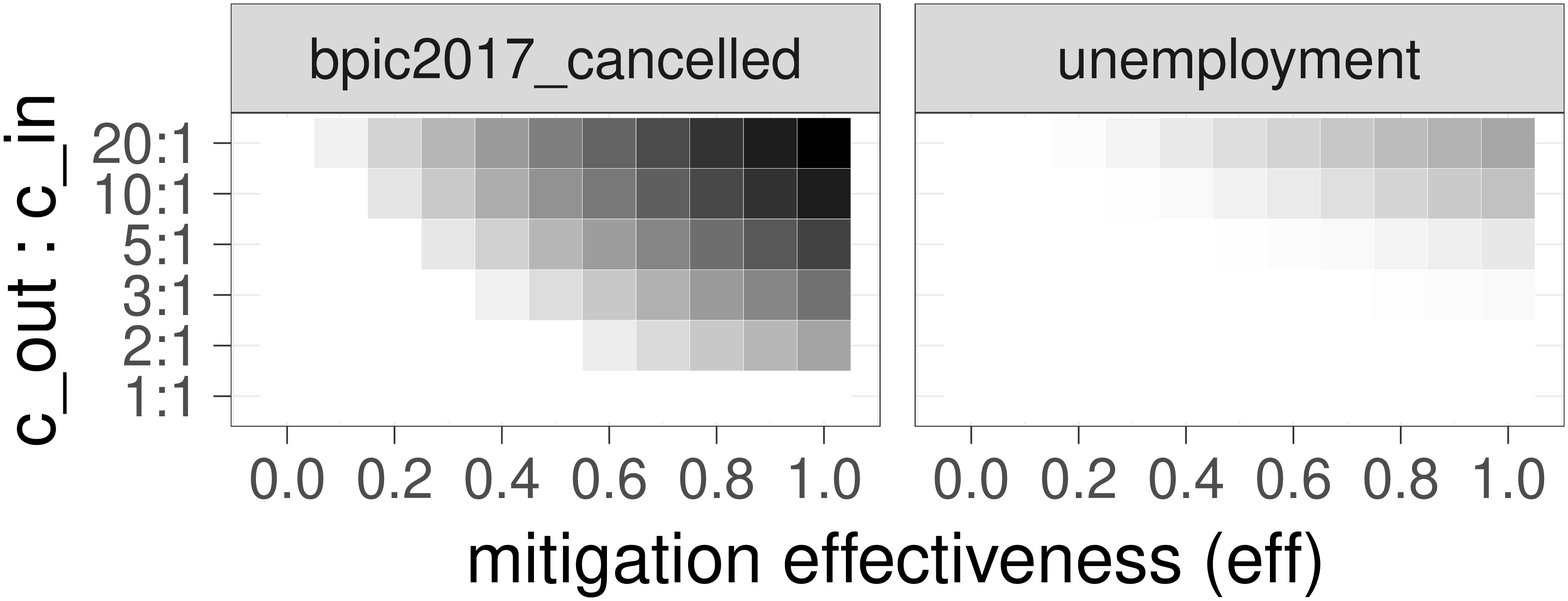}}
        \hspace{0.05cm}
        \subfloat[Varying $c_\mathit{com}(\sigma, L)$\label{fig:results_compensation_lgbm_selected}]{\includegraphics[width=0.5\linewidth]{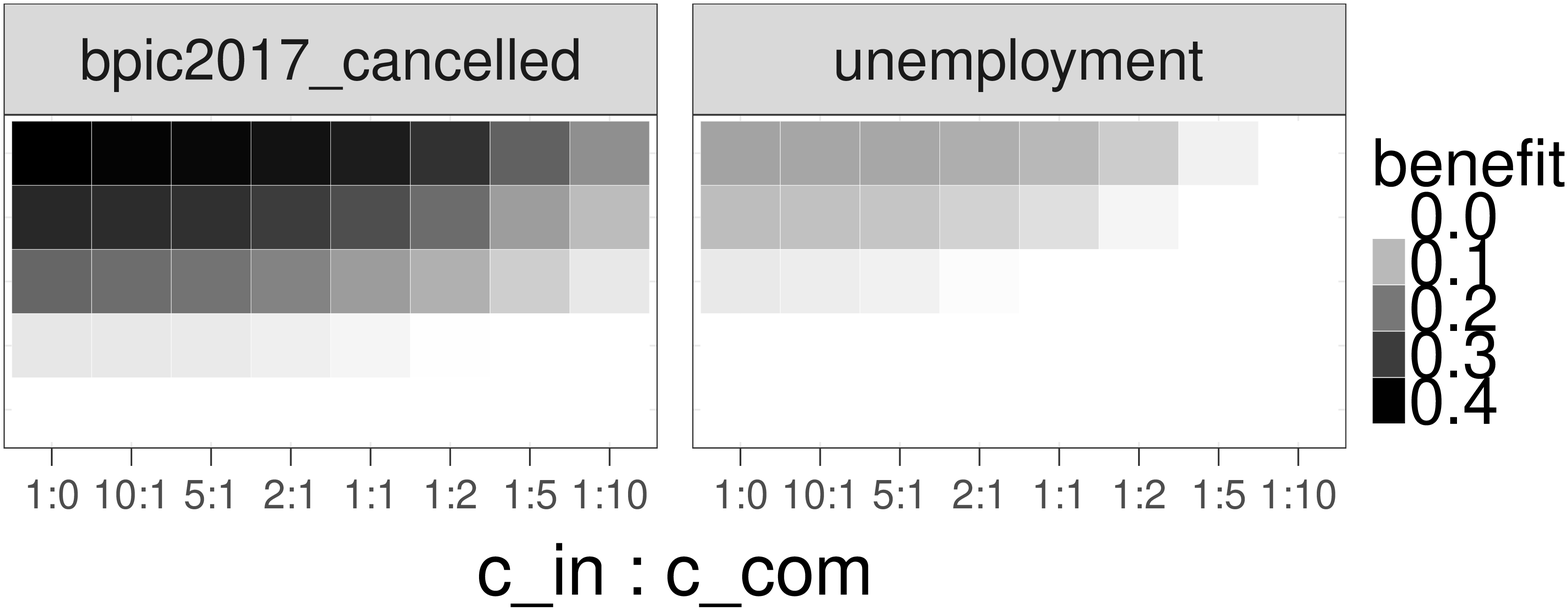}}
    \vspace{0.505\baselineskip}
    \caption{Benefit with different cost model configurations
    }
    \vspace{-0.2cm}
\end{figure}

\figurename~\ref{fig:results_effectiveness_const_lgbm_selected} shows the benefit of having an alarm system compared to not having it for different (constant) mitigation effectiveness values. As the results are similar for logs with similar class ratios, hereinafter, we only show the results for one log from each of the groups: \emph{bpic2017\_cancelled} (balanced classes) and \emph{unemployment} (imbalanced classes). As expected, the benefit increases both with higher $\mathit{eff}(k,\sigma,L)$ and with higher $c_\mathit{out}(\sigma,L):c_\mathit{in}(k,\sigma,L)$ ratio. For \emph{bpic2017\_cancelled}, the alarm system yields a benefit when $c_\mathit{out}(\sigma,L):c_\mathit{in}(k,\sigma,L)$ is high and $\mathit{eff}(k,\sigma,L) > 0$. Also, a benefit is always obtained when $\mathit{eff}(k,\sigma,L) > 0.5$ and $c_\mathit{out}(\sigma,L) > c_\mathit{in}(k,\sigma,L)$. In the case of \emph{unemployment}, the average benefits are smaller, since there are fewer cases with undesired outcome and, therefore, the number of cases where $c_\mathit{out}$ can be prevented by alarming is lower. In this case, a benefit is obtained when both $\mathit{eff}(k,\sigma,L)$ and $c_\mathit{out}(\sigma,L):c_\mathit{in}(k,\sigma,L)$ are high.
We conducted analogous experiments with linear effectiveness decay, varying the maximum possible effectiveness (at the start of the case), which confirmed that the observed patterns remain the same. We have empirically confirmed our theoretical finding (Section~\ref{sec:roi}) that $\mathit{eff}c_\mathit{out} > c_\mathit{in}$ is a necessary condition to obtain a benefit from using an alarm system, and have shown that a benefit is in practice also obtained under this condition when an optimized alarming threshold is used (cf.~RQ2).

Similarly, the benefit of the alarm system is plotted in \figurename~\ref{fig:results_compensation_lgbm_selected} across different ratios of $c_\mathit{out}(\sigma,L):c_\mathit{in}(k,\sigma,L)$ and $c_\mathit{in}(k,\sigma,L):c_\mathit{com}(\sigma,L)$. We observe that when $c_\mathit{com}(\sigma,L)$ is high, the benefit decreases due to false alarms. For \emph{bpic2017\_cancelled}, a benefit is obtained almost always, except when $c_\mathit{out}(\sigma,L) : c_\mathit{in}(k,\sigma,L)$ is low (e.g., 2:1) and $c_\mathit{com}(\sigma,L)$ is high (i.e., higher than $c_\mathit{in}(k,\sigma,L)$). For \emph{unemployment}, a benefit is obtained with fewer cost model configurations, e.g., when $c_\mathit{out}(\sigma,L):c_\mathit{in}(k,\sigma,L) = 5:1$ and $c_\mathit{com}(\sigma,L)$ is smaller than $c_\mathit{in}(k,\sigma,L)$.
We conducted analogous experiments with linearly increasing cost of intervention, varying the maximum possible cost (at the end of the case), which confirmed that the patterns described above remain the same. To answer RQ3, we have empirically confirmed that the alarm system achieves a benefit as discussed in Section~\ref{sec:roi} in case the cost of the undesired outcome is sufficiently higher than the cost of the intervention and/or the cost of the intervention is sufficiently higher than the cost of compensation.

%% file: related_work.tex
\section{Related Work}
\label{sec:related}
The problem of cost-sensitive training of machine learning models has received significant attention.
For example, Elkan~\cite{elkan2001foundations} analyzes the notion of misclassification cost and defines conditions under which a misclassification cost matrix is reasonable. Turney~\cite{turney2002types} examines a broader range of costs in the context of inductive concept learning. This latter study introduces the notion of cost of intervention, which we include in our proposed cost model. These approaches, however, do not take into account the specific costs that arise in prescriptive process monitoring.

Predictive and prescriptive process monitoring are related to Early Classification of Time Series (ECTS), which aims at classifying a (partial) time series as early as possible, while achieving high classification accuracy~\cite{xing2012early}.
To the best of our knowledge, works~\cite{mori2017early,dachraoui2015early,tavenard2016cost} are the only ECTS methods trying to balance
accuracy-related and earliness-related costs.
However, these approaches assume that predicting a positive class early has the same effect on the cost function as predicting a negative class early, which is not the case in typical business process monitoring scenarios, where earliness matters only when an undesired outcome is predicted.
Works~\cite{metzger2017predictive,di2016clustering} focus on alarm-based prescriptive process monitoring, but only allow alarms to be raised when a given state of the process is reached. This moment might potentially be late to mitigate the consequences, which would have been possible if the alarm was raised earlier. Furthermore, our approach does not require an explicit modelling of the process states. Last but not least, they rely on
a fixed-threshold alarming mechanism provided by process owners, as opposed to our empirical thresholding approach.
Gr\"oger et al.~\cite{groger2014prescriptive} is an existing approach to provide recommendations, but it misses
the two core elements of our proposed prescriptive process monitoring framework, i.e., cost models and earliness.

%% file: threats.tex
\section{Limitations and Future Work}
\label{sec:threats}

While the scenarios discussed in Boxes~1-3 show that the proposed framework is versatile enough to cover a variety of cases, the current version of the framework relies on two main assumptions. First, it assumes that an alarm always triggers an intervention, thus ignoring that a process worker might in some cases decide not to or be unable to intervene. Additionally, the current version of the framework considers each case in isolation, omitting the overall workload of the process workers, which in reality is an important factor for determining the number of alarms that can be acted upon. This limitation can be lifted by, e.g., combining the alarm system with~\cite{CONFORTI20151}, which proposes a recommender system that optimizes suggestions in case of concurrent process executions. A second limitation of the framework is that only one possible type of intervention is envisaged. This assumption can be lifted by extending the framework so that the cost of an intervention can vary depending on the specific action suggested by a recommender system.

Next to these limitations, we acknowledge the importance of further investigation on the applicability of the framework in practice.
In particular, in the future, we aim at collaborating with companies and institutions to study whether process stakeholders are able to define the costs in a natural and simple way. Also, we plan to further investigate the consequences of incorrect and/or imprecise instantiations of the cost models.
Furthermore, the current evaluation is limited to measuring the benefit of the alarm system in an offline manner, while a more thorough evaluation would consist in deploying the alarming mechanism in a real organization and making an end-to-end comparison of the costs before and after the deployment of the alarm system. However, this is a difficult task for two main reasons. First, companies need to be willing to let the technique really influence the process executions. Second, the end-to-end effectiveness analysis cannot be conducted without coupling the alarm system with a recommender system: if the system raises proper alarms, but inappropriate interventions are taken, the system would still be ineffective.
Another avenue for future work is to extend the framework with active learning methods in order to incrementally tune the alarming mechanism based on feedback about the relevance of the alarms and the effectiveness of the interventions.

%% file: conclusion.tex
\section{Conclusion}
\label{sec:conclusion}

This paper outlined an alarm-based prescriptive process monitoring framework that extends existing predictive process monitoring approaches with the concepts of alarms, interventions, compensations, and mitigation effects.
The framework incorporates a cost model to analyze the tradeoffs between the cost of intervention, the benefit of mitigating or preventing undesired outcomes, and the cost of compensating for unnecessary interventions induced by false alarms. The cost model allows one to estimate the benefits of deploying a prescriptive process monitoring system for the purposes of return on investment analysis.
Additionally, the framework incorporates a technique to optimize the alarm generation mechanism with respect to a given configuration of the cost model and a given event log.
An empirical evaluation on real-life logs showed the benefits of applying this optimization versus a baseline where a fixed likelihood score threshold is used to generate alarms, as considered in previous work in the field.